\documentclass[conference]{IEEEtran}
\IEEEoverridecommandlockouts
\usepackage{cite}
\usepackage{amsmath,amssymb,amsfonts}
\usepackage{algorithmic}
\usepackage{graphicx}
\usepackage{textcomp}
\usepackage{xcolor}
\usepackage{array}
\usepackage{mdwmath}
\usepackage{mdwtab}
\usepackage{eqparbox}
\usepackage{url}
\usepackage{hyperref}
\usepackage{tabularx}
\usepackage{subcaption}
\hypersetup{citecolor=Green}
\usepackage{mathrsfs}
\usepackage{rsfso}
\usepackage{multirow}
\usepackage{amssymb}
\usepackage{lipsum}
\usepackage{multirow, booktabs}
\usepackage{fancyhdr}
\usepackage{tabularx} 
\usepackage{subfloat}
\def\BibTeX{{\rm B\kern-.05em{\sc i\kern-.025em b}\kern-.08em
    T\kern-.1667em\lower.7ex\hbox{E}\kern-.125emX}}

\makeatletter
 \let\old@ps@headings\ps@headings
 \let\old@ps@IEEEtitlepagestyle\ps@IEEEtitlepagestyle
 \def\confheader#1{%
 
 \def\ps@IEEEtitlepagestyle{%
 \old@ps@IEEEtitlepagestyle%
 \def\@oddhead{\strut\hfill#1\hfill\strut}%
 \def\@evenhead{\strut\hfill#1\hfill\strut}%
 }%
 \ps@headings%
 }
 \makeatother

\begin{document}

\title{Location Agnostic Source-Free Domain Adaptive Learning to Predict Solar Power Generation
\thanks{}
}

\author{
\IEEEauthorblockN{Md Shazid Islam\IEEEauthorrefmark{1}, A S M Jahid Hasan \IEEEauthorrefmark{2}, Md Saydur Rahman\IEEEauthorrefmark{1}, Jubair Yusuf,\\ Md Saiful Islam Sajol\IEEEauthorrefmark{3}, Farhana Akter Tumpa\IEEEauthorrefmark{4}}

\IEEEauthorblockA{\IEEEauthorrefmark{1}University of California Riverside,California, USA,\IEEEauthorrefmark{2}North South University, Dhaka, Bangladesh,\\\IEEEauthorrefmark{3}Louisiana State University, Louisiana, USA,\IEEEauthorrefmark{4}Ahsanullah University of Science and Technology, Dhaka, Bangladesh\\Email: misla048@ucr.edu, jahid.hasan12@northsouth.edu, mrahm054@ucr.edu,\\jyusuf177@gmail.com, msajol1@lsu.edu, Tumpafarhanaakter@gmail.com}
}

% \confheader{
% \begin{flushleft}
% \textit{\textbf{$\begin{array}{l}\text {2023 IEEE International Conference on Energy Technologies for Future Grids}\\\text {Wollongong, Australia, December 3-6, 2023}\end{array}$}}    
% \end{flushleft}
% }

\maketitle

% \IEEEpubid{\makebox[\columnwidth]{\textbf{978-1-6654-7164-0/23/\$31.00~\copyright~2023 IEEE \hfill}} \hspace{\columnsep}\makebox[\columnwidth]{ }}

\begin{abstract}
The prediction of solar power generation is a challenging task due to its dependence on climatic characteristics that exhibit spatial and temporal variability. The performance of a prediction model may vary across different places due to changes in data distribution, resulting in a model that works well in one region but not in others. Furthermore, as a consequence of global warming, there is a notable acceleration in the alteration of weather patterns on an annual basis. This phenomenon introduces the potential for diminished efficacy of existing models, even within the same geographical region, as time progresses. In this paper, a domain adaptive deep learning-based framework is proposed to estimate solar power generation using weather features that can solve the aforementioned challenges. A feed-forward deep convolutional network model is trained for a known location dataset in a supervised manner and utilized to predict the solar power of an unknown location later. This adaptive data-driven approach exhibits notable advantages in terms of computing speed, storage efficiency, and its ability to improve outcomes in scenarios where state-of-the-art non-adaptive methods fail. Our method has shown an improvement of $10.47 \%$, $7.44 \%$, $5.11\%$ in solar power prediction accuracy compared to best performing non-adaptive method for California (CA), Florida (FL) and New York (NY), respectively. 

\end{abstract}


\section{Introduction}
The fast population growth along with advancement in the world economy has brought about an increase in energy usage everywhere. The US Energy Administration (EIA) estimates that by 2050 there will be more than a 50\% increment in energy usage worldwide \cite{b1}. However, relying on fossil fuels fully to meet this increased demand will cause more Greenhouse Gas (GHG) emissions as well as faster depletion of useful resources. Renewable energy sources offer the best solution to this problem as they are abundant and clean. So, renewable energy is being integrated into the grid more in both developed and developing countries alike. According to the International Energy Agency (IEA), by 2025 the combined energy generation from solar and wind will surpass the generation from coal \cite{b2}. \\

However, the problem with renewable energy generation is that they are highly dependent on the weather variables. If the values of the weather variables are known, they can be utilized to estimate the renewable generation through different statistical \cite{d0,power3,power4} or machine learning approaches \cite{d1,power5,power6,power7}. Moreover, the weather variables are highly location-dependent and vary substantially based on the climatic region of the area of interest. Unfortunately, the availability and reliability of weather data are major concerns in utilizing weather variables for any applications all over the world. In particular, the credibility of weather data in developing countries are not as sufficient as the data in developed countries to train and test the machine learning models. As mentioned earlier, developing countries are also inclining towards higher renewable energy integration into the grid. Thus, the task of solar generation estimation is imperative in these countries. However, for each location of interest, the model needs to be trained individually with a weather dataset of its own. Otherwise, using a model trained on one dataset of one location to estimate the renewable generation of some other location will introduce erroneous results due to high bias towards the source dataset. This can cause difficulties in predicting renewable generation when multiple locations of different climatic regions are of interests. Developing independent models for different locations can solve this problem but this will result in increasing computational cost, and leading to a time-consuming and storage-inefficient solution. Domain adaptation can be a good alternative to address this issue by improving the model fidelity with lower data availability. Through the use of domain adaptive methods, we can apply machine learning algorithms on one dataset to train the model and use that trained model on a different dataset to get the estimations that has similar accuracy while trained with its own features. Therefore, domain adaptation can be used to train the model on the data of one climatic region and then use that to estimate the generation of another climatic region. So, this prediction methodology becomes location agnostic while being computationally efficient, meaning a reduction in time and memory storage. In this paper, domain adaptation is implemented for solar generation estimate of different regions that have wide variations in their weather profiles. The contributions of this paper are as follows: 

\begin{itemize}
    \item Proposing a domain adaptive approach to predict solar power generation from weather features. Compared to the best performing non-adative method, our approach shows an improvement of $10.47 \%$, $7.44 \%$, $5.11\%$ for California, Floria and New York, respectively in terms of accuracy.

    \item Reducing computational cost and running time by effective feature selection and adaptation.

    \item Making the proposed method storage efficient by assuming source-free constraint which means source data is not available during adaptation.
\end{itemize}

\section{Related Works}
Many machine-learning methods have been utilized for solar energy estimation in recent years. Several traditional statistical and machine learning approaches have been compared in \cite{b3} for solar Photovoltaic (PV) generation prediction taking into account various factors that affect solar generation. The results showed that Convolutional Neural Network (CNN) and its hybrid form, sometimes with ensemble arrangements, mostly provide the best outcome. An improved Long Short Term Memory (LSTM) network with autoencoder structure (Auto-LSTM) for mid-term (24-48h) prediction of PV generation has been proposed in \cite{b4}. This network is compared with a regular Artificial Neural Network (ANN), an LSTM network, and a deep belief network (DBN). The results showed that Auto LSTM has the best performance closely followed by DBN. Here, data from 21 PV sites have been used from different locations in Germany. However, each of the sites has been trained, validated, and tested individually with its data. An improved LSTM network is proposed with an adaptive hyperparameter by making some adjustments \cite{b5}. Time learning weights (TLW), Fusion activation function (FAF), Momentum resistance method with weight estimate (MRWE), and Learning factor adaptation (LFA) have been added to the network to manage the high dimensional data, gradient disappearance, global convergence, and faster computation, respectively. Though this method shows good performance, the PV generation data is from a single PV plant located in a specific region. A more robust short-term forecasting (24h) method is proposed in \cite{b6}. The authors identified 4 intraday fluctuations through the k-medoids method and performed numerical weather prediction by using similarity points among them. The joint distribution of selected weather variables is also correctly aligned using transfer component analysis (TCA) and prediction is done through SVR. This method outperformed other methods such as SVM, GBR, ANN, ELM, etc. Though they use domain adaptive analysis, the analysis is done on the same PV station within just one climatic region. The authors in \cite{b7} discuss a type of Transfer Learning (TL) technique that trains a model with a source dataset and fine-tunes the parameters using the target dataset. Here they apply attention mechanism to assign weights to different input variables for each time step, while Dilated CNN and BiLSTM sub-components extract the spatial and temporal features of the data, respectively. Though the TL strategy is applied to two different sites of a PV station, they both are located within the same climatic region as they are from the same PV station. In \cite{b8} author addresses the challenge of forecasting wind power production for new wind farms by introducing a novel cluster-based Source Domain Adaptation (MSDA) approach that leverages existing wind farm data to improve prediction accuracy, yielding a 20.63\% reduction in regression error compared to conventional instance-based MSDA methods. The author introduces an online domain adaptive learning approach for solar power forecasting in \cite{b9}, which dynamically adjusts its model structure to accommodate changing weather conditions, enabling effective tracking of data distribution changes and reliable prediction results without relying on test data labels. Additionally, a plethora of research papers have explored the domain adaptation model in recent years, with a focus on advancing its effectiveness in the context of renewable energy sources. These endeavors aim to not only enhance the model's performance but also to contribute to the refinement of its capabilities in addressing the unique challenges posed by renewable energy domains \cite{b10,b11,b12,b13,b14,d0}. However, all of these works focus on one climatic region instead of focusing on the ability of domain adaptation to address the variability in weather features due to spatial variation. This work tries to address this research gap through the use of a source-free domain-adaptive deep learning framework where the solar PV production can be estimated for a region by deploying the model that has been developed by making use of data from another region. Both the regions are significantly different in terms of their seasonal behaviors and geographically far apart.

\section{Methodology}
\begin{figure*}
\centering
\includegraphics[width=\textwidth]{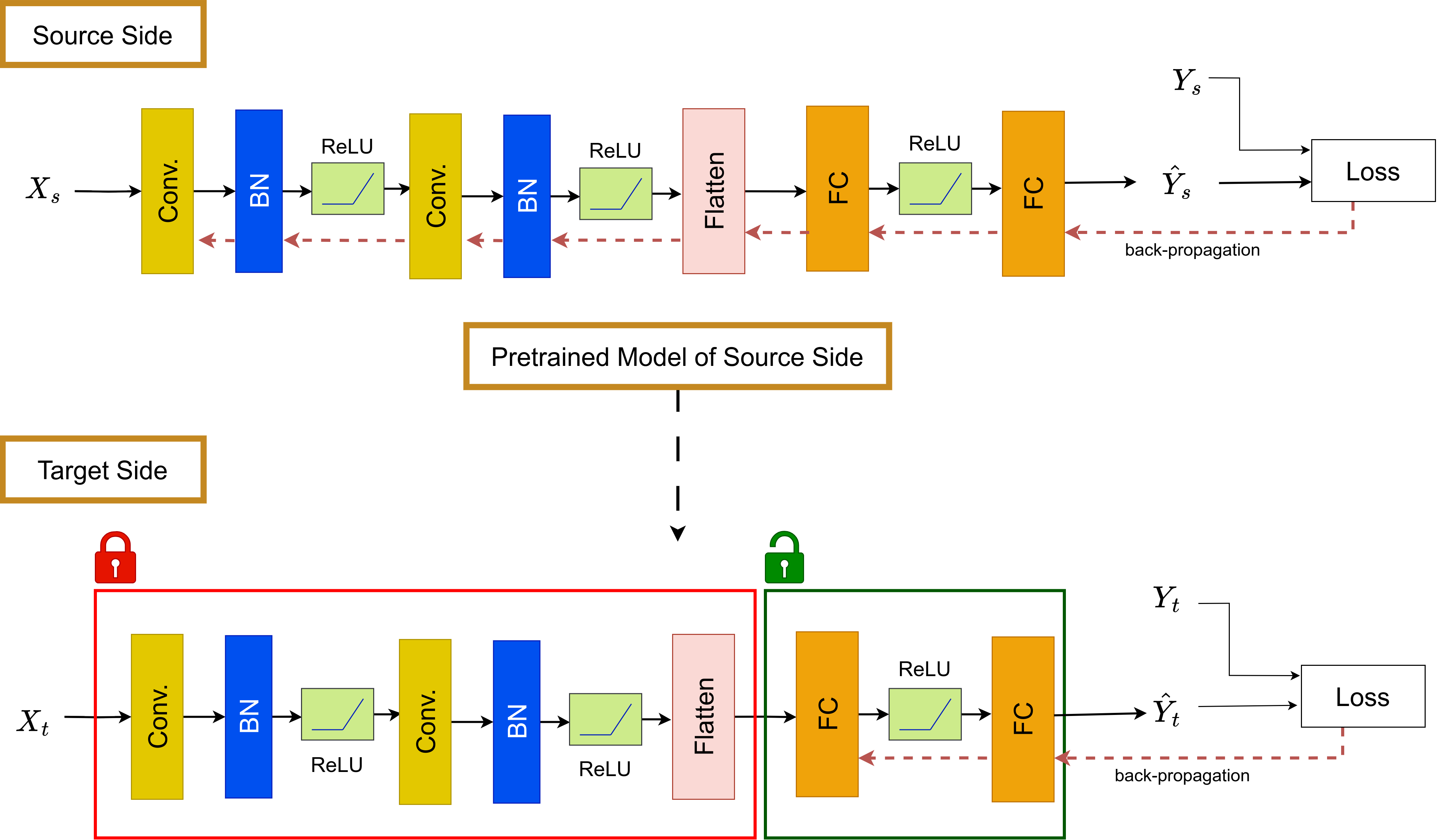}
\caption{Overview of the methodology is shown in the figure. At first a model is trained from the scratch on the source side using the source data $(X_s, Y_s)$. Then the pretrained model of the source side is transferred to the target side. On the target side, the weight of the only last two layers (FC layers) of the model is adapted using target data $(X_t, Y_t)$. The rest of the network weights are kept as same as the pre-trained model.  }
\label{overview}
\end{figure*}

We propose a method to predict the solar power generation of one location based on its meteorological characteristics using only the pre-trained model of another location. Hence we have two sets of datasets that we are referring to as source domain and target domain, respectively. The source domain involves the domain where we train a model in a fully supervised manner. When the training on the source domain is complete, only the pre-trained model is passed to the target domain. To make the process storage efficient, we assume that data of the source domain is not available on the target side. On the target domain, the pre-trained source model is adapted just by updating a few layers of the pre-trained model which makes the overall procedure faster and reduces computational cost. Hence the two main steps of our work are training a model on the source domain and adaptation of a pre-trained model on the target domain. Fig. \ref{overview} shows the total workflow of our approach.

% \subsection{Problem Statement}
\begin{figure}
\centering
\includegraphics[width=\columnwidth]{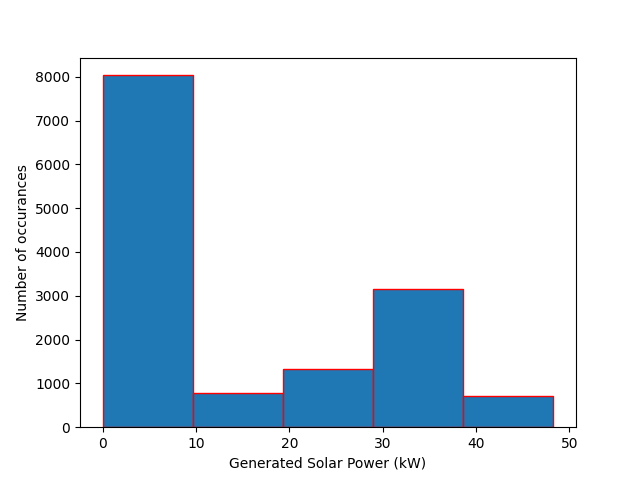}
\caption{Histogram on solar power generation shown in five bins. The first bin (low power) has highest number of frequency. Because out of 24 hrs of a day around 12 hrs (6 pm to 6 am), the sunlight is not present. Hence no solar power is generated in that period which makes the highest number of frequency in the first bin.}
\label{hist}
\end{figure}

\subsection{Training Model on Source Domain}
We shall train a feed-forward deep convolutional neural network using the source data. It is well acknowledged that neural networks can be effectively trained to address two types of tasks, namely classification and regression. We formulate our problem as a classification problem. We shall divide the range of solar power generation into $N_{c}$ number of bins (here $N_{c} = 5$). Fig. \ref{hist} shows the histogram of generated solar power of the source domain. The source data is split into training, validation and testing data.

A deep neural network of Fig. \ref{overview} comprising convolutional layers (Conv.), batch normalization layers (BN) and fully connected layers (FC) is used for training. The output layer is a FC layer which has $N_{c}$ number of nodes in order to facilitate the classification task. Input $X_{s}$ is the weather features of source domain and the output of the network is $\hat{Y}_{s}$ which is guided by the cross-entropy (CE) loss \cite{ce_loss} using ground truth of source side $Y_{s}$.

\subsection{Adaptation on Target Domain}

On the target side, we assume only the pre-trained weights of the source side are available and the source data () are not available. The objective is to update the weights of the pre-trained source model based on the data available on the target side. It is possible to update all the weights of the pre-trained network. However, it will increase the computational cost. To reduce the computational cost, only the last two layers (FC layers) of the model is updated because the last layers have the most influence on the output. Let $X_{t}$ and $Y_{t}$ are the weather features and ground truth on the target side, respectively. The output of the final FC layer is normalized by the softmax function. The normalized output ($\hat{p}_i$) can be expressed by

\begin{equation}
\hat{p}_i=\frac{e^{\hat{y}_i}}{\sum_{j=1}^{N_{c}} e^{\hat{y}_j}}
\end{equation}

where $i = \{1,2, \cdot \cdot \cdot, N_{c} \}$ and $\hat{y}_i$ is the network output on the  $i-th$ class.

Cross-entropy (CE) loss function is used to update the last two layers of the model which can be expressed by: 

\begin{equation}
\text { Loss }=-\sum_{i=1}^{ N_{c} } y_i \cdot \log \hat{p}_i
\end{equation}

\begin{table}[b]
    \caption{Comparison of accuracies between different methods in predicting solar power generation using source data for California (CA), Florida (FL) and New York (NY).}
    \begin{center}
    % c|c|
    \begin{tabular}{|c|c|c|c|}
    \hline \textbf{Classifier Method} & \textbf{CA} & \textbf{FL} & \textbf{NY} \\
    \hline Adaboost & $76.65 \%$ & $68.58 \%$ & $67.75 \%$ \\
    \hline Gradient Boosting & $78.99 \%$ & $73.14 \%$ & $70.66 \%$ \\
    \hline Random Forest& $76.71 \%$ & $68.78 \%$ & $67.92 \%$ \\
    \hline Deep Neural Network& \textbf{$81.02 \%$} & \textbf{$80.99 \%$} & \textbf{$80.59 \%$} \\
    \hline

        \end{tabular}
    
    \label{table:source}
    \end{center}
\end{table}

\begin{figure}[t]
\centering
    \subfloat[]{{\includegraphics[width= 0.8\columnwidth]{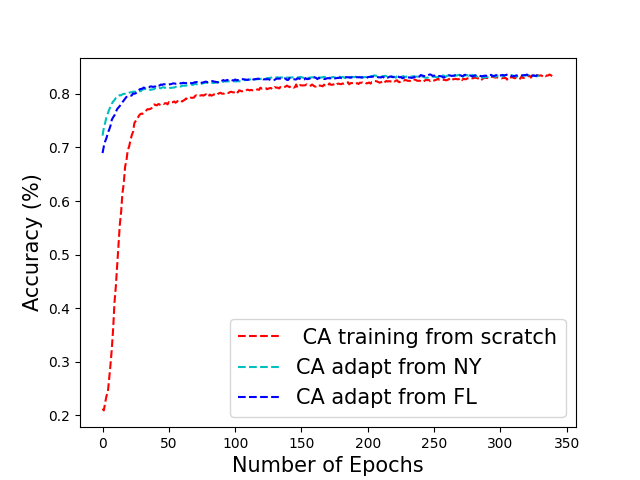}} \label{CA}}
    \vfill
    \subfloat[]{{\includegraphics[width= 0.8\columnwidth]{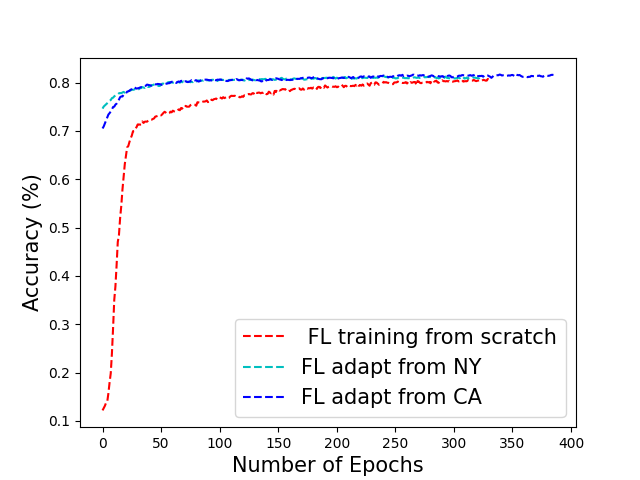}}\label{FL}}
    \vfill
    \subfloat[]{{\includegraphics[width=0.8\columnwidth]{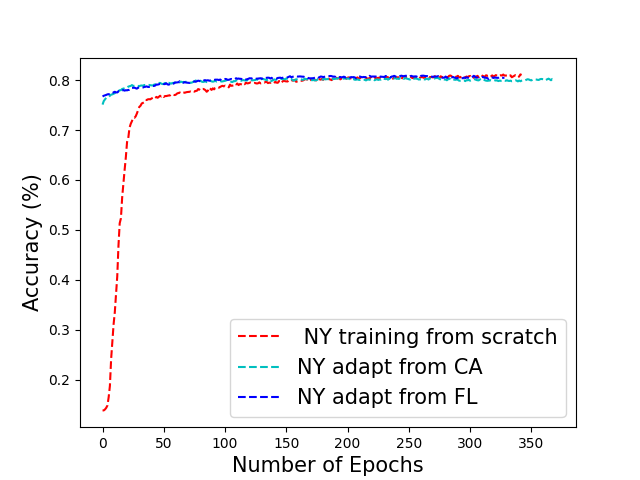}}\label{NY}}
    % \hfill
    \caption{Comparing the temporal performance between training from scratch and adaptation from a pre-trained model. In (a), (b) and (c) the target domains are CA, FL and NY, respectively. The graphs show that with domain adaptation, the network completes the training (reaches the saturation of accuracy) faster.    }\label{time}
\end{figure}

\begin{table*}[!ht]
  \caption{Comparing the performance in terms of percentage accuracies for Deep Learning Network on different locations before and after adaptation. }
  \small\centering
 
    \begin{tabular}
    % {|c|c|c|c|c|c|c|}
    {|c| c c c| c c c| c c c|}
    \hline \multirow{3}{*}{\textbf{$\begin{array}{l}\text { Source } \\
    \text { Domain }\end{array}$}} & \multicolumn{9}{c|}{ \textbf{Target Domain} } \\
    \cline { 2 - 10 } & \multicolumn{3}{c|}{\textbf{\textit{ CA}} } & \multicolumn{3}{c|}{\textbf{\textit{FL}}} & \multicolumn{3}{c|}{\textbf{\textit{NY}}} \\
    \cline { 2 - 10 } & \textbf{\textit{w/o adapt.}} & \textbf{\textit{w. adapt.}} & \textbf{\textit{diff.}} & \textbf{\textit{w/o adapt.}} & \textbf{\textit{w. adapt.}} & \textbf{\textit{diff.}}  & \textbf{\textit{w/o adapt.}} & \textbf{\textit{w. adapt.}} & \textbf{\textit{diff.}} \\
    \hline $\mathrm{CA}$ & \multicolumn{3}{c|}{$\mathrm{N} / \mathrm{A}$} & 69.43 $\%$ & 81.56 $\%$ & 12.13 $\%$ & 72.68 $\%$ & 79.94 $\%$ & 7.26 $\%$\\
    \hline $\mathrm{FL}$ & 64.09 $\%$ & 82.99  $\%$ & 18.90 $\%$ & \multicolumn{3}{c|}{$\mathrm{N} / \mathrm{A}$} & 74.88 $\%$ & 79.99 $\%$ & 5.11 $\%$\\
    \hline NY & 72.77 $\%$ & 83.24 $\%$ & 10.47 $\%$ & 74.06 $\%$ & 81.50 $\%$ & 7.44 $\%$ & \multicolumn{3}{c|}{ N/A } \\
    \hline
    \end{tabular}

\label{table:adapt}
\end{table*}

\begin{table*}[!ht]
  \caption{Comparing the performance in terms of F-1 score for Deep Learning Network on different locations before and after adaptation. }
  \small\centering
 
    \begin{tabular}
    % {|c|c|c|c|c|c|c|}
    {|c| c c c| c c c| c c c|}
    \hline \multirow{3}{*}{\textbf{$\begin{array}{l}\text { Source } \\
    \text { Domain }\end{array}$}} & \multicolumn{9}{c|}{ \textbf{Target Domain} } \\
    \cline { 2 - 10 } & \multicolumn{3}{c|}{\textbf{\textit{ CA}} } & \multicolumn{3}{c|}{\textbf{\textit{FL}}} & \multicolumn{3}{c|}{\textbf{\textit{NY}}} \\
    \cline { 2 - 10 } & \textbf{\textit{w/o adapt.}} & \textbf{\textit{w. adapt.}} & \textbf{\textit{diff.}} & \textbf{\textit{w/o adapt.}} & \textbf{\textit{w. adapt.}} & \textbf{\textit{diff.}}  & \textbf{\textit{w/o adapt.}} & \textbf{\textit{w. adapt.}} & \textbf{\textit{diff.}} \\
    \hline $\mathrm{CA}$ & \multicolumn{3}{c|}{$\mathrm{N} / \mathrm{A}$} & 81.95 $\%$ & 89.84 $\%$ & 7.89 $\%$ & 84.17 $\%$ & 88.85 $\%$ &  4.68$\%$\\
    \hline $\mathrm{FL}$ & 78.11 $\%$ & 90.70  $\%$ &  12.59 $\%$ & \multicolumn{3}{c|}{$\mathrm{N} / \mathrm{A}$} & 85.63 $\%$ & 88.88 $\%$ & 3.25  $\%$\\
    \hline NY & 84.23 $\%$ & 90.85 $\%$ & 6.62 $\%$ & 85.09 $\%$ & 89.81 $\%$ &  4.72$\%$ & \multicolumn{3}{c|}{ N/A } \\
    \hline
    \end{tabular}

\label{table:adapt2}
\end{table*}

\section{Experiment and Results}
\subsection{Dataset}
Weather data is employed as features to predict the solar generation. The solar data is retrieved from \cite{c1} and the meteorological data is collected from \cite{c2}. Synthetic solar data is available for different states that can be utilized to get a realistic representation of solar production for the given co-ordinates. This dataset is generated for 6000 simulated PV plants by using the sub-hour irradiance algorithm \cite{c3}. It consists of solar PV generation data for any specific kW capacity for 5-minute intervals for the year 2006. Solar production data is retrieved for these three states: California (CA), Florida (FL) and New York (NY). Validating the algorithm's effectiveness on these datasets will demonstrate the value of this work and the necessity of it because these regions are geographically distinct and have different climates. This solar data is only representative of year 2006 and the weather data is of that year as well. Hence, data leakage concern is out of scope for this synthetic solar dataset. However, the weather data of those co-ordinates available for that year is of 30 minutes and includes the following nine features: direct normal irradiance (DNI), direct horizontal irradiance (DHI), global horizontal irradiance (GHI), dew point, temperature, pressure, relative humidity, wind diretion, wind speed, surface albedo. Solar production data are averaged to produce 30-minute interval data in order to match the temporal resolution of meteorological data. This procedure generates 17,520 samples at 30-minute intervals, or 1 year's worth of dataset for each location where weather data are the designated features used for predicting solar production in the domain adaptation algorithm. The random forest algorithm \cite{c4} is widely used for any classification or regression task to identify the key parameters and help in eliminating irrelevant features. Hence, this algorithm is applied to find out the impactful features for predicting solar power and only a handful number of features (DNI,DHI,GHI, Temperature, Wind Direction, and Wind Speed) are utilized to accomplish the prediction task.

\subsection{Results}
In this sub-section, different types of analyses are presented on our proposed method. Firstly, the performance of the deep learning model trained on the source data will be compared against other ensembling machine learning  approaches. Secondly, the quantitative performance of the models for adaptation and non-adaptation scenarios will be evaluated as well. Finally, the impact of domain adaptation in terms of computational cost and running time will be investigated.

\subsubsection{Performance of Training Source Model from Scratch  }
We train the deep learning model using source data in order to get the pretrained model. Our implementation is executed by the popular deep learning framework Pytorch \cite{pytorch}. We use the learning rate $10^{-4}$, batch size $1000$, and popular optimization technique ADAM \cite{adam} optimizer. We compare the performance of deep learning model with some standard ensembling machine learning techniques such as Adaboost Classifier \cite{ab} , Gradient Boosting Classifier \cite{gbr}, Random Forest Classifier \cite{rfr}. TABLE \ref{table:source} illustrates that Deep Neural Network Classifier outperforms all other ensembling machine learning methods. Deep learning method improves the result at least $2.03 \%$, $ 7.85 \%$ and $9.93 \%$ for states California (CA), Florida (FL), and New York (NY), respectively. Hence choosing a deep neural network in our task was the most reasonable choice. If the hyper-parameters of the deep learning model are fine-tuned, we may achieve slightly better accuracy.

\subsubsection{Performance of Adaptation on Accuracy}

We take the pretrained model which is trained on the data of one location and adapt the pretrained model to the data of rest of the locations. In TABLE \ref{table:adapt} and \ref{table:adapt2}, we demonstrate how the results improve after adaptation in terms of accuracy and F1-score, respectively. For example, considering CA as the source domain, we found $12.13 \%$ and $7.26 \%$ improvement in accuracy, $7.89 \%$ and $4.68 \%$ improvement in F-1 score  after adaptation considering FL and NY as the target domain, respectively. Results obtained considering FL and NY as the source domain are also shown in TABLE \ref{table:adapt} . \\

\subsubsection{Performance on Running Time}
Instead of training from scratch for every domain, our proposed method uses a pretrained model from a source domain and adapts the  model for new target domains. As the model is already trained on source data, it takes less time and computation to adapt in new domain compared to training from scratch. This will be particularly helpful for large datasets and real-time very short-term solar prediction task. Moreover, lower computational cost can contribute to many aspects of solar integration such as real-time monitoring, optimization based on real-time solar estimations, etc. Fig. \ref{time} illustrates that it takes less number of epochs if adaptation on pretrained model is deployed instead of training from scratch. For example, in Fig. \ref{CA} if CA is considered as  the target domain, adaptation method is about 3 times faster than training from scratch.

\begin{table}[h]
  \caption{Comparing the performance between updating the network fully and partially during adaptation. Here partial update means updating only the last two layers of the network.}
\label{table:full_partial}

    \begin{tabular}{|c|c|c|c|c|}
    \hline \textbf{Source-Target} & \textbf{update} & \textbf{$\begin{array}{l}\text { Acc. } \\
    \end{array}$} & \textbf{F1- Sc.} & \textbf{$\begin{array}{l}\text { Avg. } \\
    \text { Speed } \\
    \text { (its/sec) }\end{array}$} \\
    \hline CA - FL & Full & 82.08 \% & 90.16 \% & 72.83\\
    \hline CA - FL & Partial & 81.56 \% & 89.84 \% & 119.68 \\
    \hline CA - NY & Full & 80.45 \% & 89.16 \% & 73.22 \\
    \hline CA - NY & Partial & 79.94 \% & 88.85 \%& 119.16 \\
    \hline
    \end{tabular}

\end{table}

\subsection{Ablation Studies}

\subsubsection{Performance comparison between updating full network and partial network}

In our work, we only update the last two layers (FC layers) of the network while working on the target domain instead on updating the whole network. In TABLE \ref{table:full_partial} , we compare the performance between updating the full network and the partial network considering CA as source domain and FL, NY as the target domain . It is evident that updating the whole network during adaption increases accuracy for FL and NY by $0.48 \%$ and $0.51 \%$, respectively. However, updating the full network decreases the speed of adaptation by $39.14 \%$ and $38.39 \%$, respectively. Hence there is a trade-off between adaptation speed and accuracy. Given the little impact on accuracy relative to the significant drop in speed, it is evident that employing a partial update of the network is a more appropriate approach when dealing with larger datasets and requiring a prompt response. Hence, this domain adaptation approach with partial network updates is more suitable in real-world scenarios. 

\begin{table}
    \caption{Comparison of accuracies between with and without effective feature selection}
    \label{feature}
    \begin{center}
        \begin{tabular}{|c|c|c|}
        \hline \textbf{Location} & \textbf{$\begin{array}{c}\text { with feature } \\
        \text { selection }\end{array}$} & \textbf{$\begin{array}{c}\text { without feature }\\
        \text { selection }\end{array}$} \\
        \hline CA & $81.02 \%$ & $79.93 \%$ \\
        \hline FL & $80.99 \%$ & $78.45 \%$ \\
        \hline NY & $80.59 \%$ & $80.39 \%$ \\
        \hline
        \end{tabular}

    \end{center}
    
\end{table}

\subsubsection{Performance comparison between with and without effective feature selection.}

In the original dataset a number of weather features are available. Only some of the important features identified by the random forest regressor are utilized  for predicting solar power generation. TABLE \ref{feature} shows the impact of selecting effective features. We observe that the use of effective features improves the results by $1.09 \%$, $1.64 \%$ and $0.20 \%$ for CA,  FL and NY, respectively. Effective feature selection helps the deep learning network to learn the patterns from the relevant features in a better way. It not only improves the results but also reduces the computational costs as the redundant features are discarded.

\section{Conclusion}

In this paper, a deep neural network-based adaptive technique is developed for the solar power prediction task from weather features. Our proposed method outperforms the non-adaptive approach in terms of accuracy. In addition, this approach does not use the source data which makes the approach storage efficient. Furthermore, the proposed adaption strategy is characterized by a reduced time needed compared to training from scratch, hence enhancing the time and computational efficiency. The conducted ablation studies indicate the practical feasibility of our proposed technique. In the future, this work will be extended for fully unsupervised domain adaptation to predict solar power generation.


\begin{thebibliography}{00}
\bibitem{b1}  Capuano, L.,Us energy information administration’s international energy outlook 2020. US Department of Energy: Washington DC, 7.
\bibitem{b2} Abdelilah, Y., Bahar, H., Criswell, T., Bojek, P., Briens, F. and Feuvre, P.L., Renewables 2020: Analysis and Forecast to 2025, IEA: Paris, France, 2020

\bibitem{d0} Tumpa, Farhana Akter, Md Saydur Rahman, and Md Shazid Islam. ``Utilizing Genetic Evolution to Enhance Cellular Automata for Accurate Image Edge Detection''. No. 10279. EasyChair, 2023. https://easychair.org/publications/preprint/3TK2

\bibitem{power3} A. S. M Jahid Hasan, M. S. Rahman, M. S. Islam and J. Yusuf, “Data Driven Energy Theft Localization in a Distribution Network," 2023 International Conference on Information and Communication Technology for Sustainable Development (ICICT4SD), Dhaka, Bangladesh, 2023, pp. 388-392, doi: 10.1109/ICICT4SD59951.2023.10303520.

\bibitem{power4} A. S. M. Jahid Hasan, J. Yusuf, M. S. Rahman and M. S. Islam,  “Electricity Cost Optimization for Large Loads through Energy Storage and Renewable Energy," 2023 International Conference on Information and Communication Technology for Sustainable Development (ICICT4SD), Dhaka, Bangladesh, 2023, pp. 46-50, doi: 10.1109/ICICT4SD59951.2023.10303409.

\bibitem{d1} Islam, Md Shazid and Rahman, Md Saydur and Amin, M Ashraful, ``Beat Based Realistic Dance Video Generation using Deep Learning"
, IEEE International Conference on Robotics, Automation, Artificial-intelligence and Internet-of-Things,  Dhaka,Bangladesh, 2019,pp. 43-47.
 doi: 10.1109/RAAICON48939.2019.22.

\bibitem{power5} M. S. Islam, M. S. Rahman, M. S. U. Haque, F. A.
Tumpa, M. S. Hossain, and A. A. Arabi,``Location
agnostic adaptive rain precipitation prediction using
deep learning,” in IEEE 9th International Women in
Engineering (WIE) Conference on Electrical and Com-
puter Engineering (WIECON-ECE), 2023 in press arXiv preprint arXiv:2402.01208



\bibitem{power6} M. S. Hossain, M. S. Islam, M. S. U. Haque, and
M. S. Rahman, “Gait phase classification from semg
in multiple locomotion mode using deep learning,” in
9th International Congress on Information and Com-
munication Technology, 2024, in press.


\bibitem{power7} M. S. Rahman, F.A, Tumpa, M. S. Islam, A. A. Arabi, M. S. B. Hossain, M. S. U. Haque, and
M. S. Rahman, “Comparative Evaluation of Weather Forecasting
using Machine Learning Models,” in
26th International Conference on Computer and Information Technology (ICCIT), 2023, in press. arXiv preprint arXiv:2402.01206

\bibitem{b3} Ahmed, R., Sreeram, V., Mishra, Y. and Arif, M.D. A review and evaluation of the state-of-the-art in PV solar power forecasting: Techniques and optimization. Renewable and Sustainable Energy Reviews, 2020.

\bibitem{b4} Gensler, A., Henze, J., Sick, B. and Raabe, N.," Deep Learning for solar power forecasting—An approach using AutoEncoder and LSTM Neural Networks", IEEE international conference on systems, man, and cybernetics, 2016, pp. 002858-002865.
\bibitem{b5} Chai, M., Xia, F., Hao, S., Peng, D., Cui, C. and Liu, W., 2019. PV power prediction based on LSTM with adaptive hyperparameter adjustment. Ieee Access, 7, pp.115473-115486.
\bibitem{b6} Wang, J., Yan, G., Ren, M., Xu, X., Ye, Z. and Zhu, Z, " Short term photovoltaic power prediction based on transfer learning and considering sequence uncertainty", Journal of Renewable and Sustainable Energy, 15(1),2023.
\bibitem{b7} Tang, Y., Yang, K., Zhang, S. and Zhang, Z.,  Photovoltaic power forecasting: A hybrid deep learning model incorporating transfer learning strategy. Renewable and Sustainable Energy Reviews, 162,2022, p.112473.
\bibitem{b8} Tasnim, S., Rahman, A., Oo, A.M.T. and Haque, M.E., Wind power prediction in new stations based on knowledge of existing Stations: A cluster based multi source domain adaptation approach. Knowledge-Based Systems, 145, 2018, pp.15-24.


\bibitem{b9} H. Sheng, B. Ray, K. Chen and Y. Cheng, "Solar Power Forecasting Based on Domain Adaptive Learning," in IEEE Access, vol. 8, pp. 198580-198590, 2020, doi: 10.1109/ACCESS.2020.3034100.

\bibitem{b10} X. Wang, Q. Kang, M. Zhou, S. Yao and A. Abusorrah, "Domain Adaptation Multitask Optimization," in IEEE Transactions on Cybernetics, vol. 53, no. 7, pp. 4567-4578, July 2023, doi: 10.1109/TCYB.2022.3222101.

\bibitem{b11}H. Cai and D. J. Hill, "Knowledge Transfer for Long-term Voltage Stability Assessment Between Power Grids Based on Deep Domain Adaptation Networks," IEEE PES Asia-Pacific Power and Energy Engineering Conference (APPEEC), Nanjing, China, 2020, pp. 1-5, doi: 10.1109/APPEEC48164.2020.9220527.

\bibitem{b12} Z. Tang, Y. Tang, A. Qiao, J. Liu and J. Gao, "Transfer Learning Based Photovoltaic Power Forecasting with XGBoost," Panda Forum on Power and Energy (PandaFPE), Chengdu, China, 2023, pp. 1781-1785, doi: 10.1109/PandaFPE57779.2023.10141226.

\bibitem{b13} Guariso, Giorgio, Giuseppe Nunnari, and Matteo Sangiorgio. "Multi-step solar irradiance forecasting and domain adaptation of deep neural networks." Energies 13, no. 15 ,2020.

\bibitem{b14} Zhang, Jun, Guozheng Peng, Rui Song, Shuhua Zhang, Yuanpeng Tan, Tianjiao Pu, and Jiye Wang. "Asymptotic domain adaptive detection for abnormal targets in transmission lines under complex weather conditions." CSEE Journal of Power and Energy Systems, 2023.

\bibitem{c1} [Online] "Solar Power Data for Integration Studies" Available: https://www.nrel.gov/grid/solar-power-data.html


\bibitem{c2} [Online] "Weather Data" Available: https://sam.nrel.gov/weather-data.html

\bibitem{c3} M Hummon,E Ibanez, G Brinkman, and D Lew.  "Sub-Hour Solar Data for Power System Modeling From Static Spatial Variability Analysis: Preprint". 2012. United States.https://www.osti.gov/servlets/purl/1059579.

\bibitem{c4} Breiman, “Random Forests”, Machine Learning, 45(1), 5-32, 2001

\bibitem{ab} Schapire, Robert E. "Explaining adaboost." Empirical Inference: Festschrift in Honor of Vladimir N. Vapnik. Berlin, Heidelberg: Springer Berlin Heidelberg, 2013, pp. 37--52.

\bibitem{gbr} Prettenhofer, Peter, and Gilles Louppe. "Gradient boosted regression trees in scikit-learn." PyData 2014. 2014.

\bibitem{rfr}Cootes, Tim F., et al. "Robust and accurate shape model fitting using random forest regression voting", European Conference on Computer Vision, Florence, Italy, 2012.



\bibitem{ce_loss} Zhang, Zhilu, and Mert Sabuncu. "Generalized cross entropy loss for training deep neural networks with noisy labels." Advances in neural information processing systems, 2018.

\bibitem{adam} Kingma, Diederik P., and Jimmy Ba. "Adam: A method for stochastic optimization." arXiv preprint arXiv:1412.6980 (2014).

\bibitem{pytorch} Imambi, Sagar, Kolla Bhanu Prakash, and G. R. Kanagachidambaresan. "PyTorch." Programming with TensorFlow: Solution for Edge Computing Applications (2021): pp. 87-104.







\end{thebibliography}
\end{document}